# RRT-CBF Based Motion Planning

Leonas Liu, Yingfan Zhang, Larry Zhang and Mehbi Kermanshabi

**Abstract**

*Control barrier functions (CBF) are widely explored to enforce the safety-critical constraints on nonlinear systems recently. There are many researchers incorporating the control barrier functions into path planning algorithms to find a safe path, but these methods involve huge computational complexity or unidirectional randomness, resulting in arising of run-time. When safety constraints are satisfied, searching efficiency, and searching space are sacrificed. This paper combines the novel motion planning approach using rapid exploring random trees (RRT) algorithm with model predictive control (MPC) to enforce the CBF with dynamically updating constraints to get the safety-critical resolution of trajectory which will enable the robots not to collide with both static and dynamic circle obstacles as well as other moving robots while considering the model uncertainty in process. Besides, this paper first realizes application of CBF-RRT in robot arm model for nonlinear system.*

## Introduction

Path planning plays a very important role in robotics that steers the robot from starting configuration to goal configuration while avoiding the collision with any obstacle along the path. The obstacles can be static or dynamic. In order to find an optimal path that can be utilized in real time for autonomous mobile robots with highly efficient navigation, many algorithms like Djikstra's [1], A* [2]and Rapidly exploring random trees (RRT) [3] are invented for path finding, but due to the limitation of the continuity of the configuration space, only RRT can be employed, which generates the path by connecting the randomly sampling points in configuration space. Moreover, even some path finding approaches based on heuristic estimation or potential fields have some benefits in optimality, the dynamic constraints and local minimum severely limit the available options.

Though it requires more time and memory to get more optimal outputs for RRT, it shows good performance if we incorporate it with other methods. Control barrier functions (CBF) [4-6] in Quadratic program (QP) is one of the methods used to enforce safety critical constraints in many applications, like adaptive cruise control [7], bipedal robot walking [8] and swarm robot [9]. Model predictive control (MPC) [10] is another method (MPC-CBF) [11-14] to handle the safety constraints but the adjusting of horizon heavily relies on the computational power of available computers and affects the safety prediction ability. Compared to other methods, such as Kinodynamic RRT* [15], LQR-RRT* [16], and the combination [17] limited by linear system and static obstacles, MPC-RRT [18] guaranteeing the safety but less flexibility in failing to balance the computational complexity and safety prediction ability (related to horizon), the combination of CBF, RRT and MPC can get a feasible safe way in acceptable time for both linear and nonlinear systems with dynamic obstacles in balancing the optimality and safety. Compared to CBF-RRT [19] and KBF-RRT [20], the CBF-RRT-MPC in this paper first consider the application for safe path planning for multi-robot system, extending the availability of autonomous path planning for real-life practical scenarios. The CBF-RRT is also originally applied to find the feasible path for a nonlinear robot arm system to execute objects grasping task.

*Contributions*: this paper first uses the similar method (CBF-RRT) as in paper [19] but in a simplified linear dynamic model to generate feasible safe paths for four robots' system, then use these paths as the reference path by using the MPC to avoid the collision for four robots. The optimization in MPC consider all circle obstacles and robot-obstacles. Since the other robots are "moving obstacles" to the selected robot, to simplify the case, I initially set circle obstacles static while testing CBF-RRT-MPC, then change these obstacles into dynamic ones. This is also a mixed case of static/dynamic circle obstacles and dynamic robot obstacles. The application of CBF-RRT on a nonlinear robot arm model is first investigated and achieved in this paper.

# Part I: CBF-RRT-MPC for Multi-robot System

## 1. Preliminaries
### 1.1 Dynamics
I consider a motion planning problem for a continuous time control system

$$\dot{x} = Ax + Bu, \tag{1}$$

where $x \in X \subset R^n$ is the system state and $u \in U \subset R^m$ is the control input, where $U$ is a set of admissible controls for system (1). The initial state is denoted as $x_{init} = x(t_0) \in X$ and the goal region is defined as $X_{goal} \subset X$. Obstacles are assumed to be moving with known constant velocity based on the equation

$$\dot{x}_{obs} = Cu, \tag{2}$$

where $x_{obs} \in X_{obs} \subset X$ is the system state for the obstacles. For simplicity, I just assume that all obstacles' velocities are known and constant.

### 1.2 Exponential Control Barrier Functions (ECBF)
Traditional path planning algorithms usually consider safety by checking if the segment of two nodes intersects obstacles. In order to be more safety critical, I formulate safety checking condition by wrapping the coordinate of nodes in a set $C$ defined by Control Barrier Functions (CBF). Given a continuously differentiable function $h: R^n \to R$, the set $C$ is defined as

$$C = \{x \in R^n | h(x) \geq 0\},$$
$$\partial C = \{x \in R^n | h(x) = 0\}, \tag{3}$$
$$Int(C) = \{x \in R^n | h(x) > 0\},$$

where $\partial C$ is the boundary of set $C$ and $Int(C)$ is the interior of set $C$. The set $C$ is forward invariance set if $x_0 \in C$ implies $x(t) \in C$ for $\forall t \geq 0$.

A continuously differentiable function $h(x)$ is an Exponential Control Barrier Function (ECBF) if there exist a positive coefficient $K \in R$ such that

$$\inf_{u \in U}[\dot{h}(x) + kh(x)] \geq 0 \ \forall x \in Int(C). \tag{4}$$

By satisfying (4), the system (1) is guaranteed forward invariant in set $C$.

### 1.3 Rapid-exploring Random Trees (RRT)
We use Tree set $T = (V, \varepsilon)$ to denote all trees with their vertices $v \in V \subset X$ and edges $e \in \varepsilon$. The classical RRT algorithm starts from the initial vertex $v_{init}$ in closed tree set and generates a vertex $v_r$ in state space based on sampling methods. We need to find out the vertex $v_c$ already in tree set which is closest to $v_r$ then a specific length edge $e_{new}$ from $v_c$ to $v_r$ is built up to get new vertex $v_{new}$. The new vertex $v_{new}$ and new edge $e_{new}$ will be stored in the closed tree set $T$ and the above process is repeated till the goal state is detected. Though the final path is bound to be feasible after finite tree growing, the sampling method is random and time costing. To improve the efficiency of CBF-RRT algorithm in this paper, we utilize steering

RRT in [19] to reduce the size of vertices and edges in tree set. The definition of steer is demonstrated in Figure 1.

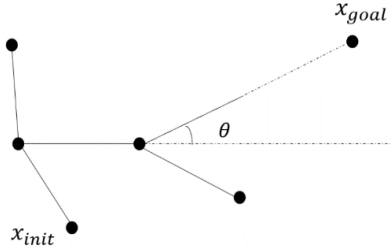

Fig. 1. Steering RRT

Different from the classical RRT, the steering RRT will not generate vertices in open state space. Instead, it just randomly picks up a vertex $v_c$ in closed tree set and stretches out a new edge $e_{new}$ from $v_c$ with steering direction denoted by angle $\theta$ from Gauss distribution to get the new vertex $v_{new}$. The following process is similar with classical RRT: store $v_{new}$ in closed tree set $T$ and repeat path searching. Compared to open state space, the closed tree usually has much smaller sampling size if searching time is not long, especially in steering RRT case, therefore steering RRT in Figure 1 usually searches faster than classical RRT algorithm.

## 2. Problem Statement and Approach

*Problem 1.* Consider a linear system in the form of (1), with initial state $x_{init} \in X_{init} \subset X \subset R^n$, where $X_{init}$ is an initial safe set, and a goal set $X_{goal}$ centered at goal point $x_{goal}$ bounded within the circle region with radius $r_{goal}$, generate feasible control inputs $u(t)$ that steer the four robots to find four paths to $X_{goal}$ while avoiding static or dynamic circle obstacles.

*Problem 2.* Generate feasible control inputs $u(t)$ that help robots track feasible paths generated from problem 1 while avoid collision with static or dynamic circle obstacles and dynamic robot obstacles.

    To approach the problems, I construct the CBF constraints by repeatedly updating the location of circle obstacles for the QP controller and use the CBF-RRT to find feasible paths. After developing the "prioritize robot" strategy to design the multi-robot system, CBF-MPC is utilized to track feasible paths as reference paths while avoid collision with static or dynamic circle obstacles and dynamic robot obstacles. This path planning algorithm is given in Section 3.

## 3. CBF-RRT-MPC

    The RRT in CBF-RRT algorithm in this paper is different from the classical RRT in terms of new node sampling and collision avoidance as discussed in Section 1.3. Based on the steering RRT process described in Figure 1, the CBF defined by the norm of Euclidean distance from robot to obstacles as (8) is embedded in the specific linear system (6) to be steering CBF-RRT as shown in Figure 2.

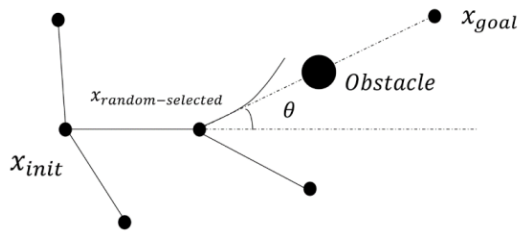

Fig. 2. Steering CBF-RRT

The reference edge length between two nodes by steering RRT is $l^r$, which is decided by control input, Gauss distributed angle and divided by 100 in steering CBF-RRT and for each sub-edge $l_k^r$ as reference, the QP controller generates the new sub-node by new sub-edge $l_k^{opt}$ in (5). The curved edge keeps away from the obstacles for safety reason.

$$x_{k+1} = x_k + l_k^{opt}, k = 0, \dots, 99. \tag{5}$$

We have several methods to design multi-robot system. Combining obstacle dynamics and all robots' dynamics into a composite system was realized in [13]. In this paper, "prioritize robot" strategy is used: the robot 1 only needs to avoid circle obstacles; robot 2 considers robot 1 together with circle obstacles as barrier; robot 3 finds feasible path based on circle obstacles, robot 1 and 2 while robot 4 navigates the free space based on circle obstacles and robot 1, 2 and 3.

The CBF-MPC helps multi-robot system track four feasible paths as reference paths constructed by all nodes and sub-nodes. The CBF-RRT-MPC algorithm can be summarized below.

---
Algorithm CBF-RRT-MPC
1: $V \leftarrow \{(x_{init}, t_{init})\}; \varepsilon \leftarrow \emptyset$                Initialization
2: $X_{goal} \subset X$
3: while $x \notin X_{goal}$, do                       Start CBF-RRT
4:     sample a vertex $v$ in $V$
5:     generate a steering angle $\theta$ based on Gauss distribution
6:     solve CBF-QP to get $u$ based on $x, \theta$ and $u_{ref}$
7:     get next $x_{new}$ based on current $u, \theta$ and $x$; update it in state set $X_{ref}$
8:     generate new vertex $v_{new}$ and new edge $e_{new}$; update them in $V, \varepsilon$
9: return whole trees set $T = (v, \varepsilon)$ and state set as $X_{ref}$
10: set control horizon $N$ and initial state $x_0$
11: while $x \notin X_{goal}$, do                      Start CBF-MPC
12:    generate a steering angle $\theta$ based on Gauss distribution
13:    solve CBF-LQR to get $u$ based on $N, x, \theta, x_{ref}$ and $u_{ref}$
14:    get next $x_{new}$ based on first step $u$ and $\theta, x$; update it in $X_{track}$
15: return state tracking set as $X_{track}$

---

## 4. Numerical Examples
### 4.1 Dynamic Systems
Consider a linear model simplified from a unicycle robot

$$\dot{x}_1 = v \cos \theta,$$

$$\dot{x}_2 = v \sin \theta, \tag{6}$$

where the state $x = [x_1, x_2]^T \in R^2$ corresponds to the location $(x_1, x_2)$ in workspace, $\theta$ is Gauss distributed angle and $v$ is velocity, also part of the control input.

## 4.2 Safety Sets
The $i$-th safety set $C_i$ is defined as
$$C_i = \{x \in R^2 : h_i(x) \geq 0\}, \tag{7}$$
where
$$h_i(x) = (x_1 - x_{1,obs,i})^2 + (x_2 - x_{2,obs,i})^2 - r_{obs,i}^2,$$
$$\dot{h}_i(x) = 2(x_1 - x_{1,obs,i})u\cos\theta + 2(x_2 - x_{2,obs,i})u\sin\theta \tag{8}$$

are defined by the norm of Euclidean distance from robot to obstacles with the center $\left(x_{1,obs,i}(t), x_{2,obs,i}(t)\right)$ and fixed radius $r_{obs,i}$.

The safe set of each robot is given as the intersection of all the safe sets for the circles
$$C_{robot} = \cap_{i=1}^{N_{obs}} C_i. \tag{9}$$

## 4.3 Sampling Distributions and CBF-RRT-MPC
The desired angle is defined below as from steering RRT
$$\theta_d = \tan^{-1}\left(\frac{x_{2goal} - x_{2c}}{x_{1goal} - x_{1c}}\right). \tag{10}$$

Then use Gauss distribution below to get $\theta$ in system (6)
$$P(\theta) = \frac{1}{\sqrt{2\pi\sigma^2}} e^{-\frac{(\theta - \theta_d)^2}{2\sigma^2}}, \tag{11}$$

where $\sigma^2$ is the variance and $\theta_d$ works as the mean. The Gauss distributed $\theta$ is embedded in both CBF-RRT and CBF-MPC.

The CBF-RRT can be generated below and solved by using Quadratic Programming (QP)
$$\min_{u \in U} \|u - u_{ref}\|^2$$
$$s.t.\ \dot{h}_i(x) + kh_i(x) \geq 0, i = 1, \dots, N_{obs} \tag{12}$$
$$\underline{u} \leq u \leq \overline{u},$$

where $u_{ref,1} = l_{k,1}^r = \frac{1}{100} v_r \cos\theta$ and $u_{ref,2} = l_{k,2}^r = \frac{1}{100} v_r \sin\theta$.

The CBF-MPC can be generated below and solved by using Linear Quadratic Regulators (LQR)
$$\min_{u \in U, x_u \in X}\left[\|x_u(N) - x_{ref}(N)\|^2 + \sum_{k=0}^{N-1}\left(\|x_u - x_{ref}\|_Q^2 + \|u - u_{ref}\|_R^2\right)\right]$$
$$s.t.\ x_1(k+1) = x_1(k) + u(k)\cos\theta$$
$$x_2(k+1) = x_2(k) + u(k)\sin\theta$$
$$x_u(0) = x_0 \tag{13}$$

$$u(k) \in U, \forall k \in [0, \ldots, N-1]$$

$$x_u(k) \in X, \forall k \in [0, \ldots, N]$$

$$\dot{h_i}(x) + kh_i(x) \geq 0, i = 1, \ldots, N_{obs}$$

$$\underline{u} \leq u \leq \overline{u}$$

$$\underline{x_u} \leq x \leq \overline{x_u},$$

where $x_{ref}$ contains the feasible path sub-nodes and $u_{ref}$ is the same as above. The control horizon here is $N = 3$.

## 5. Simulations

For the simulations, the 8 circle obstacles are in the middle region of state space and the radii of them are 0.3, 0.4 and 0.5. The radii of robot obstacles are 0.1. Velocity $v$ is bounded in $[-5,5]$ and the state $x_1, x_2$ are bounded within $[-2,8]$. I add these linear constraints in the QP with the reference velocity $v_r$ equal to 0.5 so that the robot can move in straight line when far away from obstacles and in curved line when CBF is active.

The parameter for simulation is shown in Table 1.

| Simulation | k | $\sigma^2$ | $v_r$ | $v_{obs}$ |
|---|---|---|---|---|
| 1 | 1 | 0.2 | 0.5 | 0 |
| 2 | 1 | 0.2 | 0.5 | 0 |
| 3 | 1 | 0.2 | 0.5 | 0.1 |

Table 1. Parameters for simulation

### 5.1 Single Robot CBF-RRT with Static Obstacles

The start point is green, and the goal point is black. As the simulation result shown in Figure 3, the searching process of CBF-RRT seems like the growth of tree branches (green edges) and the feasible path found is blue. This simulation result verifies the validation of the CBF-RRT algorithm for a single robot to avoid collision with several static obstacles.

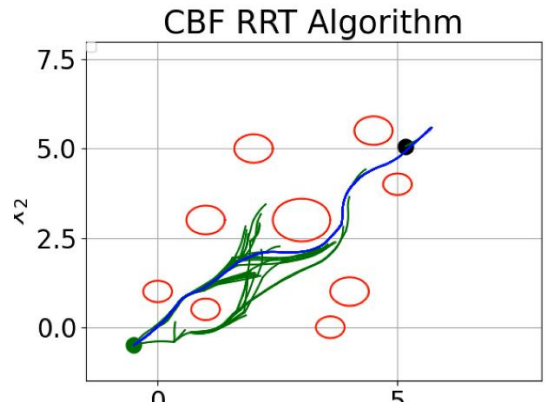

Fig. 3. CBF-RRT for single robot

## 5.2 Four Robots CBF-RRT-MPC with Static Obstacles

The start points are green, and the goal points are black. As the simulation results shown in Figure 4, 5 and 6, the CBF-RRT algorithm helps four robots find four feasible paths while avoid collision with several static obstacles. We notice that in the purple dashed square in Figure 4 and 5, the collision is about to happen for two robots when they track the feasible paths.

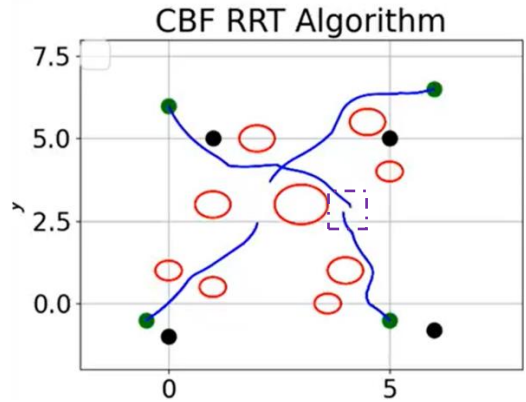

Fig. 4. CBF-RRT for four robots

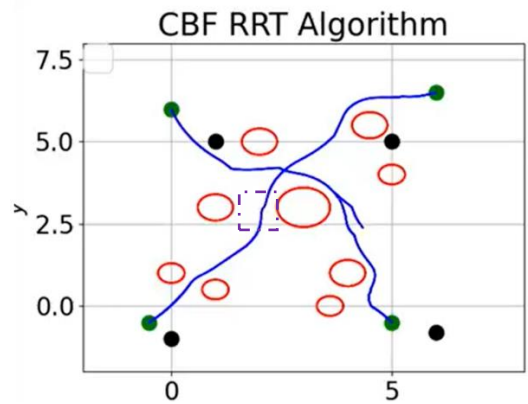

Fig. 5. CBF-RRT for four robots

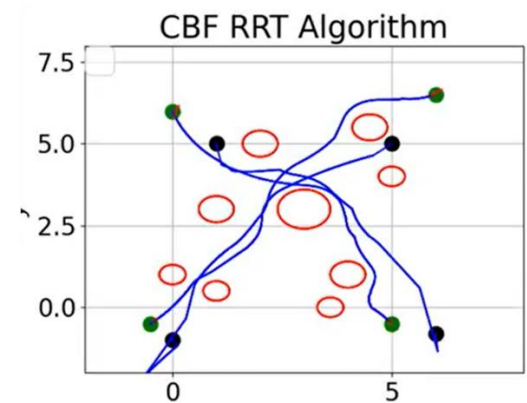

Fig. 6. CBF-RRT for four robots

To guarantee the safety, the CBF-RRT-MPC algorithm is activated, and the simulation results are shown in Figure 7, 8 and 9. Based on the simulation results below we can see the tracking paths generated by CBF-MPC are red, and CBF-RRT-MPC helps robots to avoid collision by detouring the dangerous robot obstacles around dangerous region while staying away from the static circle obstacles. Since the CBF-MPC initially requires time to process and has settling time, there will be small deviation between blue paths and red paths at the beginning.

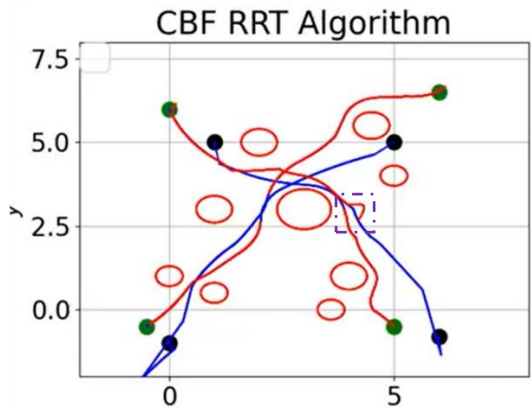

Fig. 7. CBF-RRT-MPC for four robots

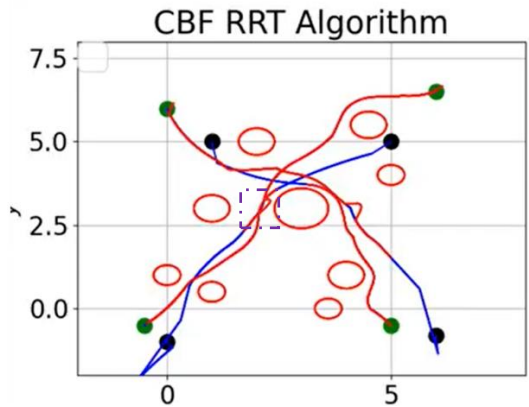

Fig. 8. CBF-RRT-MPC for four robots

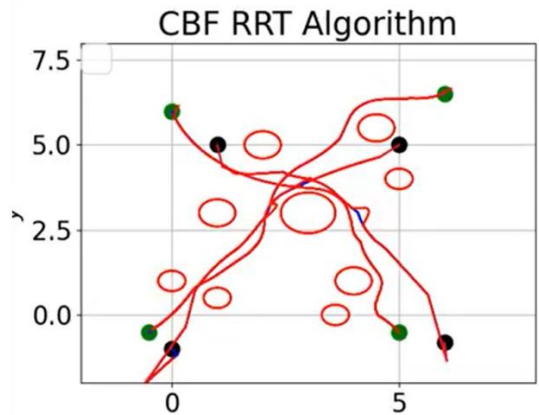

Fig. 9. CBF-RRT-MPC for four robots

## 5.3 Four Robots CBF-RRT-MPC with Dynamic Obstacles

The obstacles in above figures are all static. To make the CBF-RRT-MPC algorithm fit real practice, we enable all circle obstacles to move in specific direction with constant velocity. The simulation results are shown in Figures 10-13.

Note that in below figures we just visualize the blue tracking paths generated by CBF-MPC. The start points are green, and the goal points are red. Robots 2, 3 and 4 will detour robot obstacles and dynamic circle obstacles (the order of robots is randomly decided by CBF-RRT).

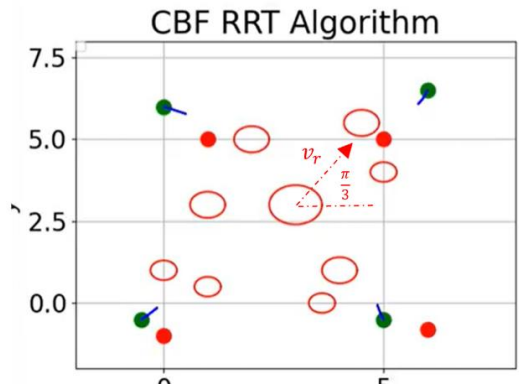

Fig. 10. CBF-MPC for four robots with dynamic obstacles

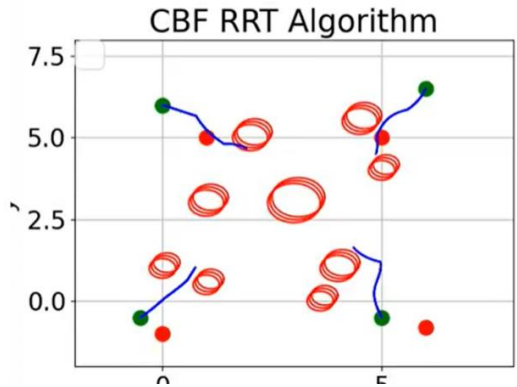

Fig. 11. CBF-MPC for four robots with dynamic obstacles

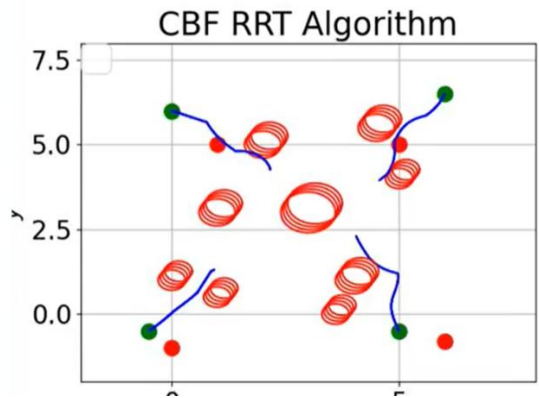

Fig. 12. CBF-MPC for four robots with dynamic obstacles

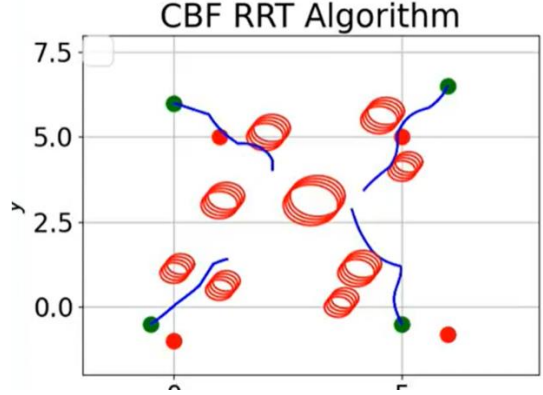

Fig. 13. CBF-MPC for four robots with dynamic obstacles

As we can see in simulation results, the CBF-MPC algorithm can work well in multi-robot system with dynamic obstacles for critical safety reason. Some paths are even close to the tangent of the dynamic obstacles but never collide with obstacles.

# Part II: CBF-RRT for Robot Arm

## 6. CBF-RRT for Single Point Robot with non-linear Dynamics

In this section, we consider utilizing CBF-RRT-based motion planning proposed by [13] on a point robot and then modify and apply it on a two-link robotic arm. The CBF-RRT algorithm is shown in Figure 14. Most problem setups are the same as [13] but we utilize a different solver and simulation environment.

**Algorithm 3** CBF-RRT

1: $\mathcal{V} \leftarrow \{(x_{init}, t_{init})\}; \mathcal{E} \leftarrow \emptyset$ ;  ▷ Initialize first vertex to initial state.
2: $\mathcal{X}_{goal} \subset \mathcal{X}$  ▷ Define goal region.
3: **while** $x \notin \mathcal{X}_{goal}$ **do**
4: $\quad v_s \leftarrow \text{VerticesSample}(\mathcal{V}, p_v)$  ▷ Sample a vertex in $\mathcal{V}$
5: $\quad v_e \leftarrow \text{StateSample}(v_s, p_{state})$  ▷ Sample the state at vertex $x_s$ (i.e. $\theta$)
6: $\quad u_{ref} \leftarrow \text{ReferenceSample}(p_{u_{ref}})$  ▷ Sample reference control if needed
7: $\quad \mathbf{u}_{traj}, \mathbf{x}_{traj}, x_{new}, t_{new} \leftarrow \text{SafeSteer}(v_e, t_h, u_{ref})$  ▷ Solve CBF QP
8: $\quad$ **if** $x_{new} \neq \emptyset$ **then**  ▷ If QP was feasible
9: $\quad\quad \mathcal{V} \leftarrow (x_{new}, t_{new}), \mathcal{E} \leftarrow \mathbf{x}_{traj}$  ▷ Update the tree
**return** $\mathcal{T} = (\mathcal{V}, \mathcal{E}), \mathbf{u}_{traj}$

**Algorithm 4** SafeSteer

Given $v_e, t_h, u_{ref}$
2: $\zeta_i \leftarrow$ i-th CBF constraint, $\forall i$;
$\mathbf{u}_{traj}, \mathbf{x}_{traj}, x_{new}, t_{new} \leftarrow \text{Integrator}(x_e, t_e, t_h, \text{QPcontroller}(x(t), x_{obs}(t), u_{ref}))$;
4: **return** $\mathbf{u}_{traj}, \mathbf{x}_{traj}, x_{new}, t_{new}$

Fig. 14. CBF-RRT-based motion planning algorithm [19]

For RRT-CBF-based motion planning in the multi-robot systems discussed in the previous section, we consider a linear model and use the velocity to control the robot. However, in this case, we consider a unicycle model for a two-wheeled differential drive robot, as shown below. ($x_1$, $x_2$) is the location and $\theta$ is the heading angle of the robot. The control input u is the bounded linear and angular velocity, i.e. $u = [(v, w)]^T$.

$$\begin{aligned} \dot{x}_1 &= v * cos(\theta) \\ \dot{x}_2 &= v * sin(\theta) \\ \dot{\theta} &= \omega \end{aligned} \quad (14)$$

We consider an environment with obstacles which are models as the circles with centroids ($x_{obs,i,1}$, $x_{obs,i,2}$) and radii $r_{obs,i}$. We denote the safety set as

$$\begin{aligned} C_i &= \{x \in R^2 : h_i(x) \geq 0\} \\ h_i(x) &= (x_1(t) - x_{obs,i,1}^2(t))^2 + (x_2(t) - x_{obs,i,2}^2(t))^2 - r_{obs,i}^2 \end{aligned} \quad (15)$$

Each obstacle has the dynamic model

$$\begin{aligned} \dot{x}_{obs,i,1} &= v_{obs,i,1} \\ \dot{x}_{obs,i,2} &= v_{obs,i,2} \end{aligned} \quad (16)$$

The $i$-th ECBF constrain is

$$\begin{aligned} \zeta_i(x) &= \pounds_f^2 h_i(x) + \pounds_g \pounds_f h_i(x) + k_1 h_i(x) + k_2 \pounds_f h_i(x) \geq 0 \\ \pounds_f h_i(x) &= 2v(x_1 - x_{obs,i,1})cos(\theta) + 2v(x_2 - x_{obs,i,2})sin(\theta) \\ &\quad - 2(x_1 - x_{obs,i,1})v_{obs,i,1} - 2(x_2 - x_{obs,i,2})v_{obs,i,2} \end{aligned} \quad (17)$$

where $k_1$ and $k_2$ are selected to ensure forward invariance.

The resultant ECBF constraint is

$$\begin{aligned} \zeta_i(x) &= 2x_1 v^2 cos^2(\theta) + 2x_2 v^2 sin^2(\theta) + (2(x_2 - x_{obs,2})cos(\theta) - 2(x_1 - x_{obs,1})sin(\theta))\omega \\ &\quad + 2v_{obs,i,1}^2 + 2v_{obs,i,2}^2 + k_1 h(x) + k_1 \pounds_f h(x) \geq 0 \end{aligned} \quad (18)$$

This constraint is only linear with respect to the angular velocity but not translational velocity so it cannot be used in the Quadratic Program. To solve this problem, we use a constant translational velocity and only control the angular velocity.

For VerticesSample (CBF-RRT algorithm line 3), we consider a discrete uniform distribution over all vertices, i.e. all vertices have equal probability to be the expanding vertex. We utilize a MATLAB function called "unidrnd" to generate the random numbers from the discrete uniform distribution specified by the number of vertices.

For State Sample (CBF-RRT algorithm line 4), given the state of the expanding vertex selected from line 3, we can compute the angle $\theta_{goal}$ heading toward the goal based on the location of the expanding and goal vertex. Then, we use a Gaussian distribution (MATLAB function "makedist" ('Normal')) and heading angle as a mean to generate the state angle $\theta_{state}$. The variance of Gaussian distribution can be adjusted depending on more or less exploration we want in the workspace.

For Safe Steer (CBF-RRT algorithm line 7), we use the resultant ECBF constraints indicated before as linear constraints and then use MATLAB function "quadprog" to solve the Quadratic Program.

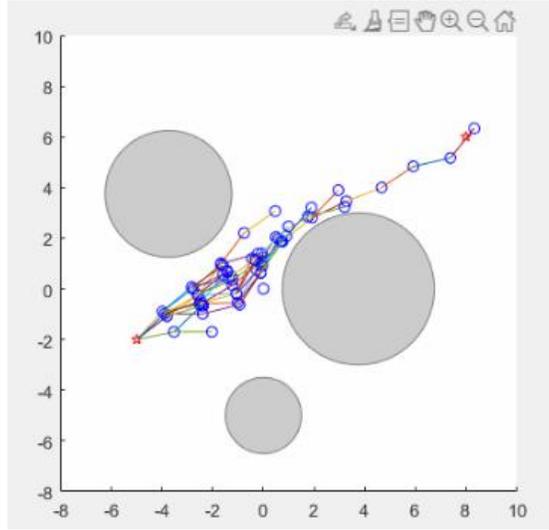

Fig. 15. CBF-RRT motion planning in a static environment

For MATLAB simulation, we consider an environment with 3 static obstacles. The start position is [-5;2] and the goal position is [10;6]. We define three obstacles with their centroids at [3.75;0], [0;5] and [-3.75;3.75] with radii 3, 1.5 and 2.5, respectively. The variance is 0.1, $k_1$ is 2 and $k_2$ is 4. The translational velocity is 1.5. The result of the simulation is shown above.

## 7. CBF RRT for Robotic Manipulators
### 7.1 Dynamic of the System

In this paper, we control the kinematic of the robot for simplicity; thus, the differential equations can be written as (19). $I$ is identity matrix and $u$ is our control input.

$$\dot{\theta} = Iu. \tag{19}$$

### 7.2 Control Barrier Functions

The most important part of this work is calculating CBFs for robot manipulators. One approach can be using control points on each link and calculating CBF constraints. This method is fast; however, safety might be compromised in some situations. In this paper, each link is over-approximated as a cylinder, and the distance between each cylinder and obstacles is our CBF constraint. In order to calculate this distance, each coordination of each obstacle will be transformed in the frame attached to the link, and the distance between the axis of the cylinder and the obstacle can be easily calculated. This method is illustrated in (20). In those equations, $^{F_j}P_W$ is the matrix that maps to the world frame to link j frame, $F_j$ is link j frame $^FX$ represents the coordination of point $X$ in $F$ frame, and $X_{obst_i}$ is set of obstacle i coordination. For developing the theory, it is assumed in each frame, the x-axis is aligned with the axis of the cylinder; therefore, the direction of distance is perpendicular to x-direction.

$$^{F_j}X_{obst_i} = {^{F_j}P_W} *^W X_{obst_i}. \tag{20}$$

This idea can be deployed for obstacles with any differentiable shape, but for simplicity, it is assumed our obstacles are spheres for the rest of the paper. If our obstacles are sphere we can construct CBF constraints like (21). $X_{c_i}, r_{obst_i}$ $n$ and $r_{link_j}$ are center of obstacle i , radius of obstacle i, number of links and radius of link j.

$$norm_{ij} = \sqrt{{}^{F_j}X_{c_i}(2)^2 + {}^{F_j}X_{c_i}(3)^2}, \tag{21}$$

$$h_{ij} = norm_{ij}^2 - (r_{obst_i} + r_{link_j})^2. \tag{22}$$

In (23), (24) and (25) $\dot{h}_{ij}$ is calculated.

$$\dot{h}_{ij} = 2 \times {}^{F_j}X_{c_i}(2)^{F_j}\dot{X}_{c_i}(2) + 2 \times {}^{F_j}X_{c_i}(3)^{F_j}\dot{X}_{c_i}(3), \tag{23}$$

$${}^{F_j}\dot{X}_{c_i} = {}^{F_j}P_W \times^W X_{c_i} + {}^{F_j}P_W \times^W X_{c_i}, \tag{24}$$

$${}^{F_j}P_W = \frac{\partial^{F_j}P_W}{\partial \theta_1}\omega_1 + \frac{\partial^{F_j}P_W}{\partial \theta_2}\omega_2 + \ldots + \frac{\partial^{F_j}P_W}{\partial \theta_n}\omega_n. \tag{25}$$

### 7.3 Activating CBF Constraints

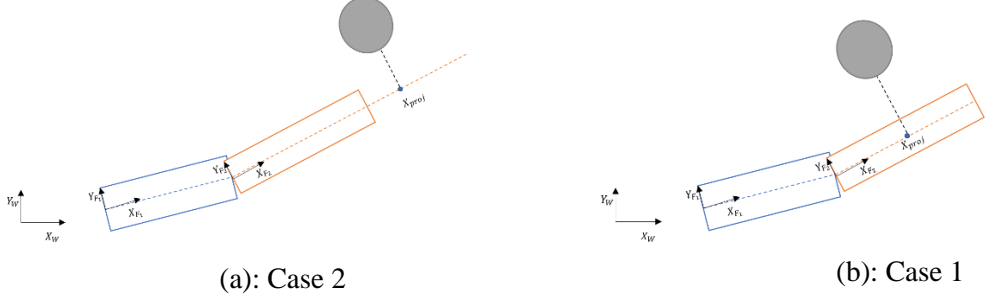

(a): Case 2      (b): Case 1

Fig. 16. Two cases for activating CBF constraints

The above equations do not consider the length of links, and it is supposed cylinders are extended to infinity. In order to tackle this issue, we need to activate each CBF constraint when the link is close to the obstacle. Thus, the distance between the obstacle and the cylinder calculated and when this distance is less than a threshold the CBF constraint is activated. So as to achieve more reliable performance, two thresholds are defined for two cases($\delta_1, \delta_2$). Suppose $X_{proj}$ is the projection of center of the obstacle on the axis of the cylinder. Case 1: When $X_{proj}$ is inside of the cylinder. Case 2: When there is an offset between $X_{proj}$ and the cylinder. These cases are demonstrated in figure 16. The reason for separating these cases is in case 1, collision with the obstacle is always possible, but in case 2, if the threshold is too large, the algorithm considers obstacles that never touch the manipulator. Therefore, the threshold for two cases should not be equal and the threshold for case 1 ($\delta_1$) larger. This part can be summarized as:

```
if case 1 then
    if dis(Cylinder, X_{c_i}) < δ_1 then
        h_{ij} ← active
    end if
end if
if case 2 then
    if dis(Cylinder, X_{c_i}) < δ_2 then
        h_{ij} ← active
    end if
end if
```

Figure 17. Case judgement

### 7.4 Quadratic Programming

For steering the robot from a vertex to the new sample point this optimization problem should be solved (for all active $h_{ij}$).

$$\min_{u} \quad (u - u_{ref})^2$$

$$s.t. \quad \dot{h}_{ij}(x_{obst_i}, \theta) + kh(x_{obst_i}, \theta) \geq 0. \tag{26}$$

$u_{ref}$ is equal to the direction of the new sampled point in configuration space, and K is a constant. In each iteration, quadratic programming is solved for a time horizon($t_h$). If the solution is infeasible while expanding a new edge, the edge is added to the tree, and then another vertex will be selected.

### 7.5 RRT

The utilized RRT variant is similar to [19] with additional enhancements in order to reach the goal with fewer iteration. Like [19] a vertex is selected to expand in a direction. However, in this paper, the sampled vertex is not entirely random, and vertexes with fewer children are more likely to select. Moreover, sometimes the direction is biased toward the destination, and other times it is completely random. In the following, state sampling and vertex sampling are explained further.

#### 7.5.1 Vertex Sampling

In this approach vertex sampling is completely random for some iteration and in other iteration one vertex with minimum number of children is chosen. $\eta_{vs}$ determines the ratio of random selection respect to biased selection.

#### 7.5.2 State Sampling

To make a balance between exploring the environment and moving toward the goal, the algorithm switches between selecting a direction randomly and biased toward the goal. Biased direction has Gaussian distribution ($\mu = direction_{goal}, \sigma^2$). $\eta_{ss}$ determines the ratio of iteration for random selection respect to biased selection.

# 8. Simulation
## 8.1 Implementation for two link manipulator
### 8.1.1 Rigid Body Transforamtion
In order to describe the coordination of obstacles in links' frame, rigid body transformation matrices are calculated in following (Links frame are demonstrated in figure 18:

$$^{F_1}P_W = R_z\theta_1), \tag{27}$$

$$^{F_2}P_W = R_z(\theta_2)T_x(-L_1)^{F_1}P_W, \tag{28}$$

Where $T, R$ are translation and rotation matrix in 3D space and all lengths are given in table 2.

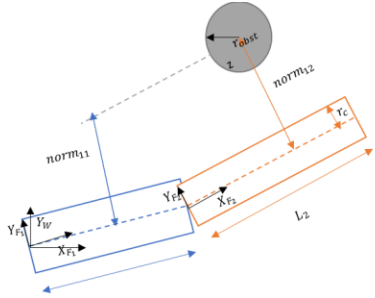

Fig. 18. Links' frame for two link manipulator

### 8.1.2 CBF Constraints
Distance between obstacle $i$ and two link manipulator is calculated in (29) and (30)

$$norm_{i1} = (y_c cos(\theta_1) - x_c sin(\theta_1))^2 sin(\theta_2)), \tag{29}$$

$$norm_{i2} = (y_c cos(\theta_1 + \theta_2) - x_c sin(\theta_1 + \theta_2) + L_1 sin(\theta_2)), \tag{30}$$

and $\dot{h}$ is given in (31), (32), (33) and (34)

$$\dot{h} = \begin{pmatrix} a_{11} & a_{12} \\ a_{21} & a_{22} \end{pmatrix} \begin{pmatrix} \dot{\theta}_1 \\ \dot{\theta}_2 \end{pmatrix}, \tag{31}$$

$$a_{11} = 2(x_c cos(\theta_1) + y_c sin(\theta_1))(y_c cos(\theta_1) - x_c sin(\theta_1)), \tag{32}$$

$$a_{12} = 0, \tag{33}$$

$$a_{21} = -2(x_c cos(\theta_1 + \theta_2) + y_c sin(\theta_1 + \theta_2))(y_c cos(\theta_1 + \theta_2) - x_c sin(\theta_1 + \theta_2) + L_1 sin(\theta_2)),$$

$$a_{22} = -2(x_c cos(\theta_1 + \theta_2) + y_c sin(\theta_1 + \theta_2) - L_1 cos(\theta_2))(y_c cos(\theta_1 + \theta_2) - x_c sin(\theta_1 + \theta_2) + L_1 sin(\theta_2)). \tag{34}$$

### 8.1.3 Result
For the simulation, the algorithm was tested in an environment with four obstacles. Obstacle located at $[5; 0]$ with radius 3, $[0; -5]$ with radius 1.5, $[-3.75; 6]$ with radius 2.5 and $[6; 6]$ with radius 2. Hyper-parameters are given in table??. For this path $\theta_{intial} = [2; 2.1]$ radiant and $\theta_{goal} = [1.35; -0.3]]$ radiant. The generated path and trees are shown in figure 19.

Table 2. Hyper-parameters for two link simulation

| $\sigma^2$ | $k$ | $t_h$ | $r_c$ | $L_1$ | $L_2$ | $\eta_{vs}$ | $\eta_{ss}$ | $\delta_1$ | $\delta_2$ |
|---|---|---|---|---|---|---|---|---|---|
| 0.4 | 2 | 0.3 | 0.3m | 3m | 3m | 3 | 3 | 5m | 0.5 m |

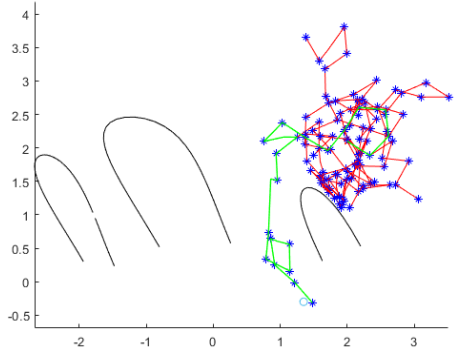

Fig. 19. Path planning in configuration space, red lines are edges; blue stars are vertices, and green lines are generated. Black curves represent obstacles.

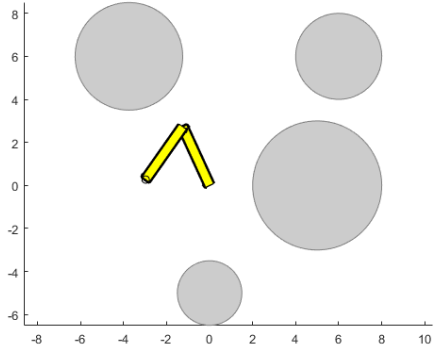

Fig. 20. Tested environment for two links manipulator

## 8.2 Implementation for Baxter Robot

In this section, the algorithm was utilized for the left arm of Baxter manipulator. Baxter has seven degrees of freedom (four for position and three for orientation. Since obstacle avoidance is our goal, the four DoF of position were controlled).

### 8.2.1 Rigid Body Transforamtion

In order to describe the coordination of obstacles in links' frame, rigid body transformation matrices are calculated in following equations (for the left arm)[21]:

$$^{BL}P_W = R_z\left(\frac{\pi}{4}\right)T_y(h)T_x(-L)T_z(-H), \tag{35}$$

$$^{BL}P_{F_1} = R_z(\theta_1)\,^{BL}P_W, \tag{36}$$

$$^{F_2}P_W = R_z(-\theta_2 - \tfrac{\pi}{2})T_x(-L_1)R_x(\tfrac{\pi}{2})^{BL}P_{F_1}, \tag{37}$$

$$^{F_3}P_W = T_z(-L_2)R_z(-\theta_3)R_x(-\tfrac{\pi}{2})^{F_2}P_W, \tag{38}$$

$$^{F_4}P_W = R_x(\tfrac{\pi}{2})T_x(-L_3)R_z(-\theta_4)^{F_3}P_W. \tag{39}$$

Where $T, R$ are translation and rotation matrix in 3D space and all lengths are given in table 3.

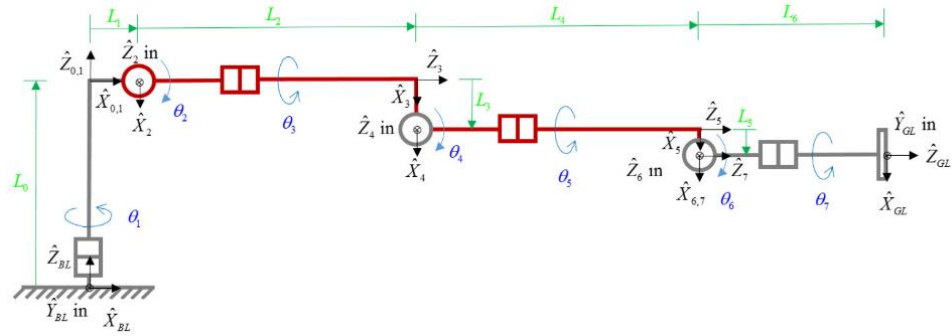

Fig. 21. Links' frames for the left arm

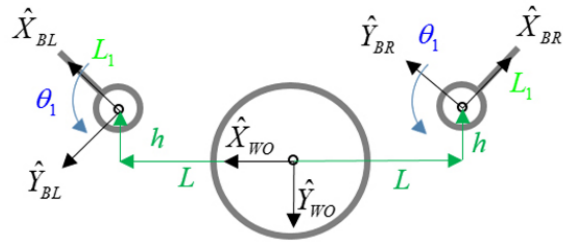

Fig. 22. Top view

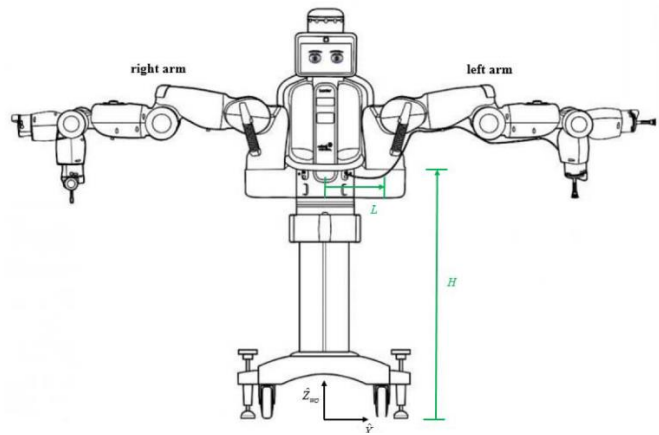

Fig. 23. Front view

Table 3. Length of Links

| Length | Value(mm) |
|---|---|
| $L$ | 278 |
| $h$ | 64 |
| $H$ | 1104 |
| $L_0$ | 270.35 |
| $L_1$ | 69.00 |
| $L_2$ | 364.35 |
| $L_3$ | 69.00 |
| $L_4$ | 374.29 |
| $L_5$ | 10.00 |
| $L_6$ | 368.30 |

### 8.2.2 Sampling in Configuration Space

To sample in 4D space of $\theta = [\theta_1; \theta_2; \theta_3; \theta_4]$ the goal biased sampling algorithm needs to modified. First, the unit direction toward the goal is calculated. Then, a random vector is generated with Gaussian distribution($\mu = 0, \sigma^2$), and the random vector's length always is equal to one.

### 8.2.3 Result

The algorithm was tested in a simple environment with single obstacle For the simulation. The obstacle located at [0.75; 0.5; 0] with radius 0.5. Other hyper-parameters are given in table 4. For this path $\theta_{intial} = [0; -\frac{\pi}{3}; 0; 0]$ and $\theta_{goal} = [-\frac{\pi}{3}; 0; 0; \frac{\pi}{2}]$. CBF functions are shown in figure 24 when they are active. Since they are all positive the algorithm steers the robot toward the goal without collision with the obstacle as you can see in figure 25.

Table 4. Hyper-parameters for Baxter simulation

| $\sigma^2$ | $k$ | $t_h$ | $r_c$ | $\eta_{vs}$ | $\eta_{ss}$ | $\delta_1$ | $\delta_2$ |
|---|---|---|---|---|---|---|---|
| 0.4 | 2 | 0.3 | 0.5m | 2 | 1 | 5m | 0.5 m |

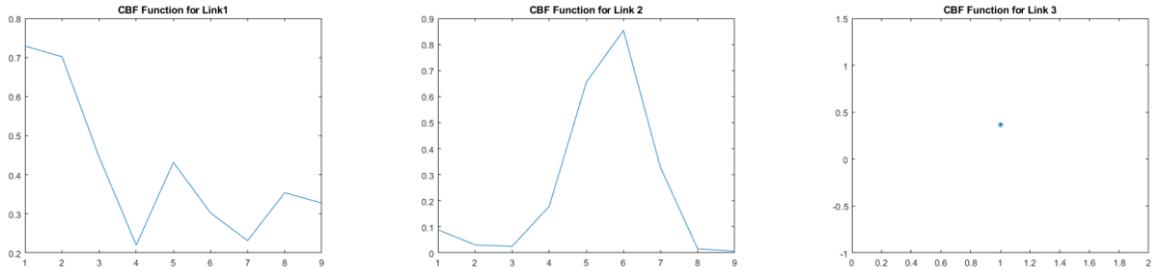

Fig. 24. CBF functions

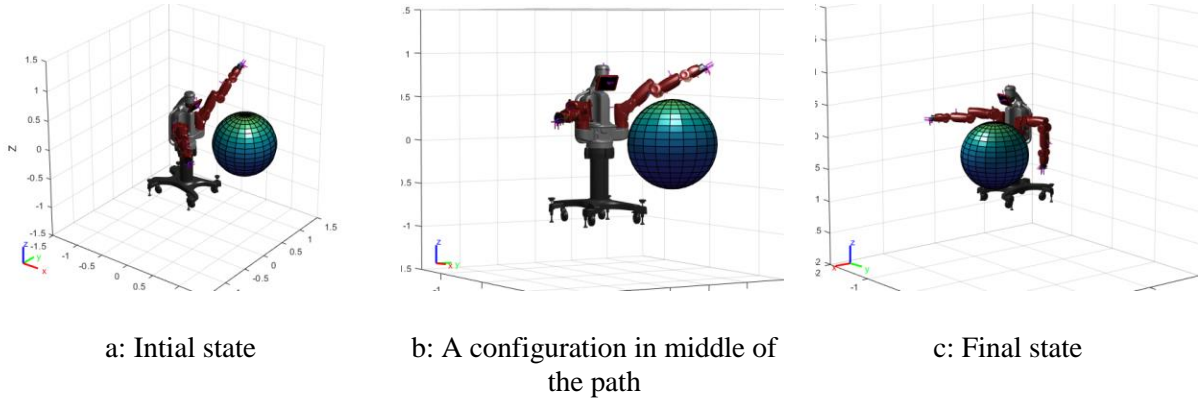

| a: Intial state | b: A configuration in middle of the path | c: Final state |

Fig. 25. Test envoironment for Baxter robot

## 9. Conclusion and Future Work

In conclusion, the CBF-RRT-MPC algorithm validly gets the safety-critical resolutions of trajectory for a linear system which will enable the robots not to collide with both static and dynamic circle obstacles as well as other moving robots while considering the model uncertainty in process. In future work, we will take place of the linear system by a unicycle nonlinear model with disturbance. Obstacles will also be considered with more complicated and unknown dynamics. To improve the searching efficiency, the algorithm should be extended and optimized to search feasible paths while consider avoiding collision in RRT. There are also several motion planning methods for multi-robot system, such as Consider Conflict Based Search (CBS) algorithm. We would like to take some investigation on CBS together with CBF to design multi-robot path planning.

With some enhancement, CBF-RRT was successfully for a robotic arm. This approach can generate safe paths without requiring or nearest neighbor collision checks. Moreover, this algorithm can handle dynamic environments as well. For future work, this algorithm can be improved by over-approximating obstacles with ellipsoids instead of spheres and using temporal logic constraints in order to handle more sophisticated constraints.

## References

[1] A note on two problems in connection with graphs, EW Dijkstra Numerische mathematik 1 (1), 269-271, 1959

[2] A formal basis for the heuristic determination of minimum cost paths, PE Hart, NJ Nilsson, B Raphael, *IEEE transactions on Systems Science and Cybernetics* 4 (2), 100-107, 1968

[3] S. M. LaValle, "Rapidly-exploring random trees: A new tool for path planning," 1998.

[4] S. Liu, W. Xiao, and C. A. Belta, "Auxiliary-variable adaptive control barrier functions for safety critical systems," in *2023 62nd IEEE Conference on Decision and Control (CDC)*, 2023, pp. 8602–8607.

[5] S. Liu, W. Xiao, and C. A. Belta, "Feasibility-guaranteed safety-critical control with applications to heterogeneous platoons," *arXiv preprint arXiv:2310.00238*, 2023.

[6] S. Liu, W. Xiao, and C. A. Belta, "Auxiliary-variable adaptive control lyapunov barrier functions for spatio-temporally constrained safety-critical applications," *arXiv preprint arXiv:2404.00881*, 2024.

[7] A. D. Ames, J. W. Grizzle, and P. Tabuada, "Control barrier function based quadratic programs with application to adaptive cruise control," in *53rd IEEE Conference on Decision and Control*, pp. 6271–6278, IEEE, 2014.

[8] S.-C. Hsu, X. Xu, and A. D. Ames, "Control barrier function based quadratic programs with application to bipedal robotic walking," in *2015 American Control Conference (ACC)*, pp. 4542–4548, IEEE, 2015

[9] U. Borrmann, L. Wang, A. D. Ames, and M. Egerstedt, "Control barrier certificates for safe swarm behavior," *IFAC-PapersOnLine*, vol. 48, no. 27, pp. 68–73, 2015

[10] Mayne, D. Q., Rawlings, J. B., Rao, C. V., and Scokaert, P. O. (2000). Constrained model predictive control: Stability and optimality. *Automatica*, 36(6):789–814.

[11] Safety-critical model predictive control with discrete-time control barrier function, J Zeng, B Zhang, K Sreenath, *2021 American Control Conference (ACC)*, 3882-3889

[12] S. Liu, J. Zeng, K. Sreenath, and C. A. Belta, "Iterative convex optimization for model predictive control with discrete-time high-order control barrier functions," in *2023 American Control Conference (ACC)*, 2023, pp. 3368–3375.

[13] S. Liu, Z. Huang, J. Zeng, K. Sreenath, and C. A. Belta, "Iterative convex optimization for safety-critical model predictive control," *arXiv preprint arXiv:2409.08300*, 2024.

[14] S. Liu and Y. Mao, "Safety-critical planning and control for dynamic obstacle avoidance using control barrier functions," *arXiv preprint arXiv:2403.19122*, 2024

[15] D. J. Webb and J. Van Den Berg, "Kinodynamic rrt*: Asymptotically optimal motion planning for robots with linear dynamics," in *2013 IEEE International Conference on Robotics and Automation*, pp. 5054–5061, IEEE, 2013.

[16] A. Perez, R. Platt, G. Konidaris, L. Kaelbling, and T. Lozano-Perez, "Lqr-rrt*: Optimal sampling-based motion planning with automatically derived extension heuristics," in *2012 IEEE International Conference on Robotics and Automation*, pp. 2537–2542, IEEE, 2012.

[17] G. Goretkin, A. Perez, R. Platt, and G. Konidaris, "Optimal sampling-based planning for linear-quadratic kinodynamic systems," in *2013 IEEE International Conference on Robotics and Automation*, pp. 2429–2436, IEEE, 2013.

[18] Rapidly Exploring Random Trees-based initialization of MPC technique designed for formations of MAVs, Zdeněk Kas, Martin Saska, Libor Přeuči, *2014 11th International Conference on Informatics in Control, Automation and Robotics (ICINCO)*

[19] Sampling-based motion planning via control barrier functions, Guang Yang, Bee Vang, Zachary Serlin, Calin Belta, Roberto Tron, *Proceedings of the 2019 3rd International Conference on Automation, Control and Robots*, 2019, 22-29

[20] Safe and Robust Motion Planning for Dynamic Robotics via Control Barrier Functions, Aniketh Manjunath, Quan Nguyen, *arXiv preprint arXiv:2011.06748*, 2020

[21] Baxter Humanoid Robot Kinematics© 2017 Dr. Bob Productions Robert L. Williams II Ph. D. williar4@ ohio. edu Mechanical Engineering Ohio University April 2017.